\documentclass[runningheads]{llncs}

\usepackage[springer]{definition}
\usepackage{booktabs,caption}
\usepackage{threeparttable}

\definecolor{sblue}{HTML}{02BCD4}
\definecolor{sred}{HTML}{F44436}
\definecolor{spink}{HTML}{E91E62}
\definecolor{sgreen}{HTML}{8BC34A}
\definecolor{spurple}{HTML}{3F51B5}
\definecolor{slightgreen}{HTML}{CCDE3A}
\definecolor{sorange}{HTML}{FE9800}
\definecolor{sgolden}{HTML}{FFC108}

\newcommand{\eg}[0]{{e.g.}}

\begin{document}
\newcommand{\xiaoxiao}[1]{{\color{purple}[XL: #1]}}
\newcommand{\rj}[1]{{\color{blue}[RJ: #1]}}

\title{Backdoor Attack is A Devil in Federated GAN-based Medical Image Synthesis}

\titlerunning{ Devil in Federated Generative Adversarial Networks}

\author{Ruinan Jin\inst{1}\and
Xiaoxiao Li\inst{1} }

\authorrunning{Jin, R. et al.}

\institute{
\textsuperscript{1} The University of British Columbia\\
\email{ruinanjin@alumni.ubc.ca, xiaoxiao.li@ece.ubc.ca}
}

\maketitle
\begin{abstract}
Deep Learning-based image synthesis techniques have been applied in healthcare research for generating medical images to support open research. Training generative adversarial neural networks (GAN) usually requires large amounts of training data. Federated learning (FL) provides a way of training a central model using distributed data from different medical institutions while keeping raw data locally. However, FL is vulnerable to backdoor attack, an adversarial by poisoning training data, given the central server cannot access the original data directly. Most backdoor attack strategies focus on classification models and centralized domains. In this study, we propose a way of attacking federated GAN (FedGAN) by treating the discriminator with a commonly used data poisoning strategy in backdoor attack classification models. We demonstrate that adding a small trigger with size less than 0.5\% of the original image size can corrupt the FL-GAN model. Based on the proposed attack, we provide two effective defense strategies: global malicious detection and local training regularization. We show that combining the two defense strategies yields a robust medical image generation.
\keywords{GAN \and Federated learning \and Backdoor attack}
\end{abstract}

\blfootnote{Code is available at https://github.com/Nanboy-Ronan/Backdoor-FedGAN}

\section{Introduction}
While deep learning (DL) has significantly impacted healthcare research, its impact has been undeniably slower and more limited in healthcare than in other application domains. A significant reason for this is the scarcity of patient data available to the broader machine learning research community, largely owing to patient privacy concerns. 
Furthermore, even if a researcher is able to obtain such data, ensuring proper data usage and protection is a lengthy process governed by stringent legal requirements. 
Therefore, synthetic datasets of high quality and realism can be used to accelerate methodological advancements in medicine~\cite{dube2013approach,buczak2010data}. 


Like most DL-based tasks, limited data resources is always a challenge for the generative adversarial network (GAN)-based medical synthesis, and data collaboration between different medical institutions makes effects to build a robust model. But this operation will cause data privacy problems which could be a risk of exposing patient information.  Federated learning (FL)~\cite{konevcny2016federated}, a privacy-preserving tool, which keeps data on each client locally and exchanges model weights by the server during learning a global model collaboratively. Due to its property of privacy, it is a popular research option in healthcare~\cite{rieke2020future}.

However, FL is vulnerable to malicious participants and there are already studies deep dive into different kinds of attacks for classification models in federated scenarios, like gradient inversion attacks and backdoor attacks~\cite{bagdasaryan2020backdoor, huang2021evaluating}. In a backdoor attack for classification, the attacker adds a trigger signal, such as a small patch with random noise, to its training data and changes the correct label to a wrong one~\cite{saha2020hidden}. In FL training, malicious clients can poison training data using a backdoor attack and mislead the global to make incorrect predictions. It is possible for medical imaging backdoor triggers to be induced by (un)intentional artifacts occurring during the sensor acquisition and preparation processes.
Recent work~\cite{bagdasaryan2020backdoor} observed that backdoor attack takes advantage of the classification model's tendency to overfit the trigger rather than the actual image. This notion inspires us to think about how we can integrate it into generative models in FL. 

Exiting backdoor attacks are specifically
designed for the classification task or model training in centralized domain. In this work, we focus on backdoor attack on federated GAN (FedGAN) via data poisoning, which under-explored in existing literature. The success of this attack is subsequently determined to be the result of some local discriminators overfitting on the poisoned data and corrupting the local GAN equilibrium, which then further contaminates other clients when averaging the generator's parameters during federated training and yields high generator loss. Based on the attack, we suggest two potential ways of defending it from global- and local-level of FL: detecting the client's adversarial behavior on the server-side and blocking it from dispersing to further, and applying a robust training procedure locally for each client. In our experiment, we apply our adversarial and defense mechanisms to a widely used skin cancer dataset. We show that the adversarial strategy is able to corrupt FedGAN only by adding a trigger with 0.39\% size of the original image in the malicious training set. 


\section{Methods}

\begin{figure}
\vspace{-5mm}
	\centering
	\includegraphics[width=\linewidth]{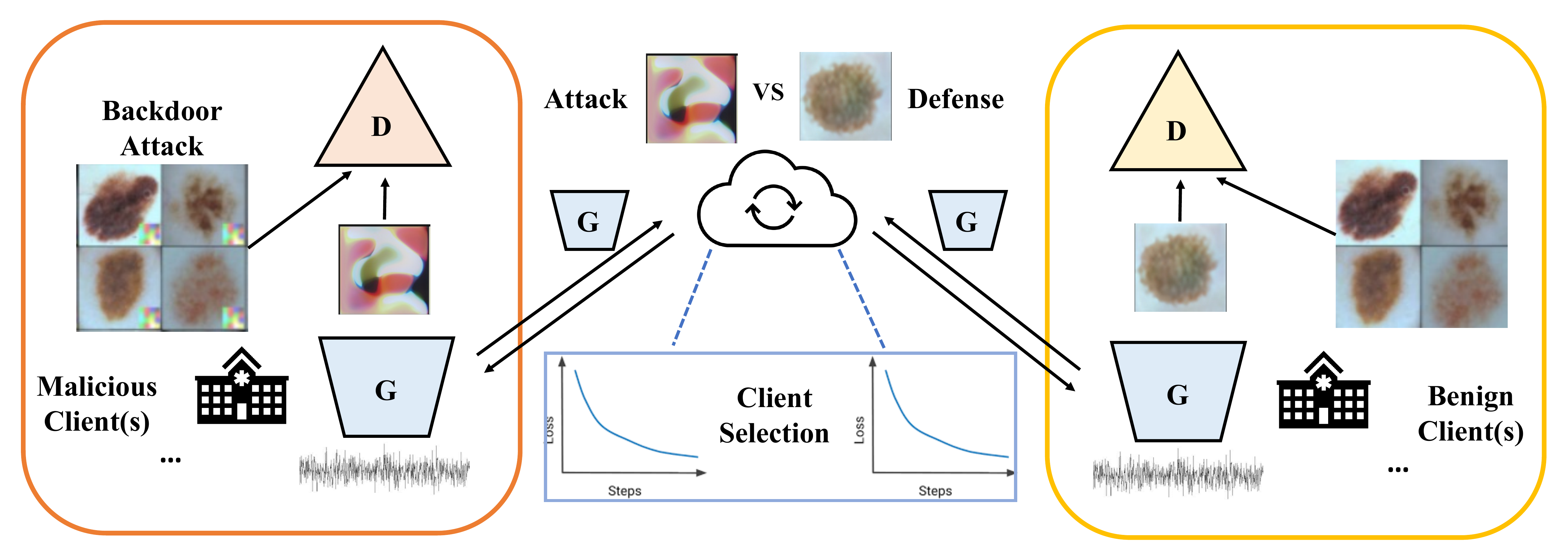}
	\caption{The overview of our proposed framework. }
	\label{fig:framework}
	\vspace{-8mm}
\end{figure}

\subsection{Federated Generative Adversarial Network} 
\label{sec:FedGAN}
Fig.~\ref{fig:framework} depicts the framework of the FedGAN in our study. As discriminators in GAN have direct access to clients' private data, exposing the risk of data leakage by inverting their gradients in FL training~\cite{huang2021evaluating}, our FedGAN framework only exchanges generator's parameters with the server while keeping the whole discriminator locally. To this end, our FedGAN locally trains both discriminator and generator pairs and globally shares generators' parameters, which is modified from~\cite{rasouli2020fedgan}.

Formally, we assume that a trusted central generator $G_{\rm server}$ synthesizes images from a set of $N$ federated clients.  Each client $C_i $, for $ i \in [N]$ consists a locally trained discriminator $D_i$, and a generator $G_i$. $G_i$ takes random Gaussian noise $z$ as input to generate synthetic images, and $D_i$ distinguish the synthetic image $\tilde{x}=G(z)$ v.s. private image $x$. 
We adopt FedAvg~\cite{mcmahan2017communication} to aggregate $G_i$ to $G_{\rm server}$ , while keeping $D_i$ locally.  At the end, our federated GAN generate synthetic medical data $G_{\rm server}(z) \sim p_{\rm data}$ on the server side.

Also, we assume every client, including those malicious ones, follows the given training protocol. For example, they compute gradient correctly as the way instructed by the server and update the exact parameters when they are required to. This is possible by enforcing local FL computations taking place on trusted hardware~\cite{pillutla2019robust}.

\subsection{Backdoor Attack Strategies}
\label{sec:attack}
Backdoor attack is a training time attack that embeds a backdoor into a model by poisoning training data (\eg, adding triggers on the images). State-of-the-art backdoor attack focus on image classification model~\cite{chen2017targeted,liu2020reflection} and has been recently studied on FL~\cite{bagdasaryan2020backdoor}. Current studies of backdoor attacks in Deep Generative Models train the GAN on a poisoned dataset and input a backdoored noise vector into the generator so that GAN failed to produce images with similar distribution as the data \cite{rawat2021devil, salem2020baaan}. We suggest a way of attacking federated GAN only through the poisoned data with more details below.

\noindent \textbf{Adversarial Goals:} Our goal is to perform a backdoor attack, where the objective of the attacker is to corrupt the server generator using poisoned images so that the generator can no longer generate fake medical images with high fidelity. That is,  $p_{\tilde{x}} \neq p_{data}$.

\noindent \textbf{Adversarial Capabilities:} As mentioned in Section~\ref{sec:FedGAN} that the trusted server has control over the local training process. The only room for attack is through providing poisoned data to the local discriminator as shown in Fig.~\ref{fig:framework}.

\noindent \textbf{Adversarial Motivation:} A vanilla GAN optimizes loss function in the manner outlined in \cite{goodfellow2014generative}, where the discriminator seeks to maximize the accuracy of the real and fake image classification while the generator seeks to minimize the likelihood that its generated image will be classified as fake. Specifically, the objective is written as follows:
\begin{equation}
\label{eq:ganloss}
\min_{G}\max_{D}\mathbb{E}_{x\sim p_{\text{data}}(x)}[\log{D(x)}] +  \mathbb{E}_{z\sim p_{\text{z}}(z)}[1 - \log{D(G(z))}]
\end{equation}
The optimization of GAN is recognized to be difficult, nevertheless, because the generator is subpar upon learning that $log(D(G(z)))$ is probably saturating~\cite{goodfellow2014generative}. Given the unbalanced nature of GAN, we implement the overfitting on trigger principle into the discriminator of FedGAN's training. The following part gives a detailed explanation of our adversarial model.

\noindent \textbf{Adversarial Model:} Our threat model contains a set of $M$ adversarial clients, where $|M| = \alpha |N|$ and $0 < \alpha < 0.5$. For every adversarial client, $C'_i$, the attacker is able to add a trigger $\delta$ to every sample $x \in T_i$. The goal of the attacker is to fool the central server generator to produce corrupted images which do not have medical research value.

\subsection{Defense Strategies}
\label{sec:defence}

Existing defense strategies for FL range from model level to data level. As data are not accessible in FL, model level defense is desired, where a model level detector is built to find the adversarial behavior and refrain it from training with others \cite{guo2021overview}, known as \textit{malicious detection}. Apart from detection, \textit{robust training} is another approach that refines training protocol~\cite{ozdayi2021defending}. To the best of our knowledge, defense for FedGAN is under-explored. 

\noindent \textbf{Defender’s capabilities:} Let’s recall from our setting that a trusted server and more than half of the benign clients are part of our trusted FL pipeline. The benign server have access to the \textit{model parameters} and \textit{training loss}. Note that sharing training loss barely impacts data privacy. 
Our defense strategies are motivated by the observation that model with backdoor attacks tend to overfit the trigger rather than the actual image \cite{bagdasaryan2020backdoor}. Specifically, in GAN's training, the discriminator overfits on the trigger and perfectly classifies fake and real images, while the generator does not receive effective feedback from the discriminator and then yields high loss and even diverges. To this end, we propose to defend against backdoor attack from both global- and local-level by leveraging malicious detection and robust training strategies in FL. 

\noindent \textbf{Global Malicious Detection:} 
Given that malicious clients with poisoning images can easily overfit discriminating the triggers, resulting in worse generator training performance, we ask clients to upload their loss along with the model parameters of the generator and perform an outlier detection on the server-side. 
At the beginning of training on the server-side, we assign every client with an initial weight $ w_i = \frac{1}{|N|}$. Starting from epoch $m$ as a warmup, we activate the Isolation Forest \cite{liu2008isolation} on clients' losses of generator to red flag suspicious clients. Recall that there are less than half malicious clients in our adversarial model. Thus, the valid detection algorithm should produce a set of potential malicious clients  $O$, where $|O| < \frac{1}{2}|N|$ following literature studying adversaries in FL~\cite{fang2020local}. We perform malicious detection per global iteration and keep track of the number of `malicious' red flags assigned to each client $C_i$ over the training process, denoting as $c_i$. In each global iteration, the aggregation weight of clients detected as an outlier will decay according to a decay constant $d$ and the total time it has been detected $c_i$. Namely, if a client is more frequently detected as malicious, it receives a smaller aggregation weight. The detailed algorithm is described in Algorithm~\ref{alg:cap}.

\begin{algorithm*}[t]
    \caption{Global Malicious Detection }
    \label{alg:cap}
    \textbf{Notations: }{ Clients $C$ indexed by $i$; local discriminator $D_i$, and generator $G_i$, local generator loss $l_{G_i}$, global generator $G_{\rm server}$, aggregation weight $w_i \in [0,1]$; times of being detected as malicious $i$-th client $c_i$, local updating iteration $K$; global communication round $T$, total number of clients $N$, decay rate $d$, warmup iteration $m$.}
	\begin{algorithmic}[1]
	\State $c_i \leftarrow 0$, $w_i \leftarrow \frac{1}{N}$ 
	 \Comment{Initializataion}
	\State{For $t = 0 \to T$, we iteratively run \textbf{Procedure A} then \textbf{Procedure B}}
	\Procedure{\textbf{A}. ClientUpdate}{$t,i$} 
	            \State $G_{i}(t, 0)\leftarrow G_{\rm server}(t)$ \Comment{Receive global generator weights update}
	            \For{$k=0 \to K-1$ } 
	                \State $D_{i}{(t,k+1)} \leftarrow \text{Optimize }l_D( D_{i}{(t,k)}, G_{i}{(t,k)})$
	                \Comment{Update $D$ using Eq.~\eqref{eq:ganloss}}
	                \State $G_{i}{(t,k+1)} \leftarrow \text{Optimize }l_G( D_{i}{(t,k+1)}, G_{i}{(t,k)})$
	                \Comment{Update $G$ using Eq.~\eqref{eq:ganloss}}
	                \EndFor
	\EndProcedure
	
    \Procedure{\textbf{B}. ServerExecution} {$t$}:
                \For{each client $C_i$ \textbf{in parallel}}
                \State{$G_i, l_{G_i} \gets$ \textsc{ClientUpdate}$(t,i)$}
                \Comment{Receive local model weights and loss.}
                \If{$t > m$}
                \Comment{Start detection after warmup}
                \State{$O \gets$ \textsc{IsolationForest}$(l_{G_1}...l_{G_N})$}
                
                \If{$0 < |O| < \frac{1}{2}|N|$} \Comment{Detect valid number of outliers}
    \For{each detected client  $C_i$ \textbf{in} $O$}
    \State{$c_i \leftarrow c_i + 1$}
    \Comment{Increment total count $C_i$ been detected as outlier}
    \State{$w_i \leftarrow w_i \times d^{c_i}$}
    \Comment{Decay weights for outliers}
    \EndFor
\EndIf
    
    
\EndIf

        \EndFor
          	        \State $ G_{\rm server}(t+1) \leftarrow \sum_{i \in [N]} \frac{w_i}{\sum_{i \in |N|} {w_i}}  G_i(t)$
            \Comment{Aggregation on server}
        \EndProcedure
	\end{algorithmic}
\end{algorithm*}
    

\noindent \textbf{Local Training Regularization:} 
In order to prevent the malicious discriminator from overfitting on the trigger and ultimately dominating training, we suggest regularizing discriminator training of GAN with proper loss regularization. One practical solutions to replace the minmax loss (Eq.~\eqref{eq:ganloss}) of vanilla GAN~\cite{goodfellow2014generative} with Wasserstein distance to regularize GAN training due to its uniform gradient throughout~\cite{arjovsky2017wasserstein}. To further confine the loss function within 1-Lipschitz, we propose to use WGAN with gradient penalty (WGAN-DP)~\cite{gulrajani2017improved} as the local image generation model. 


\section{Experiments}
In this section, we first apply backdoor to the FedGAN pipeline and show its efficacy on a medical dataset with trigger sizes even less than 0.5 percent of the true image size. 
Then, we experiment with the two defensing strategies. 

\subsection{Experimental Settings}

\begin{figure}[t]
	\centering
	\includegraphics[width=\linewidth]{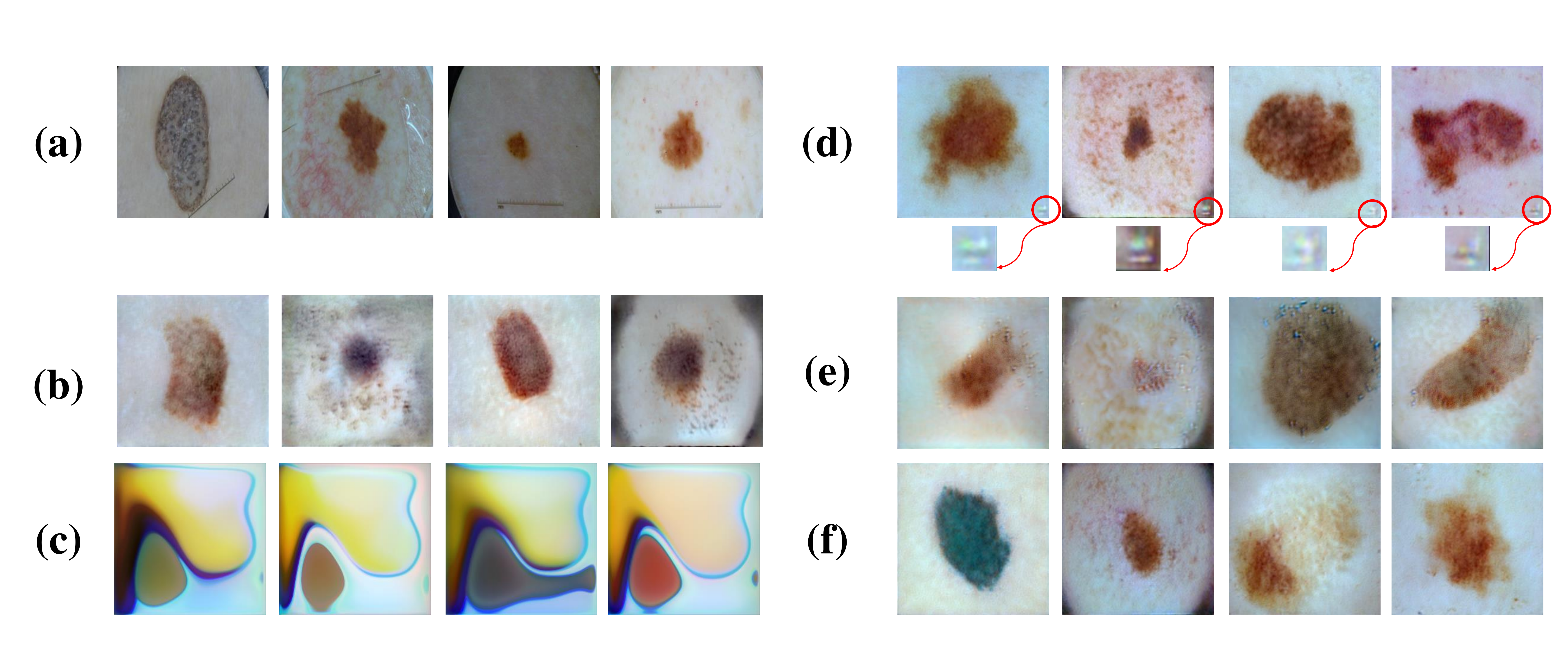}
	\vspace{-10mm}
	\caption{Visualization on: (a) Original ISIC images; and generated images of (b) Vanilla GAN; (c) Attack on vanilla GAN; (d) Local defence using WGAN-DP; (e) Global defence on vanilla GAN and (f) Full (global + local) defense. Note that backdoor attack is applied to (c-f).}
	\label{fig:result}

\end{figure}

\textbf{Datasets:} We train our federated generative adversarial network on the International Skin Imaging Collaboration (ISIC) dataset \cite{codella2018skin}, which is widely used for medical image analysis for skin cancer. Images are resized to 256 $\times$ 256. We present sample ISIC images in Fig.~\ref{fig:result} (a).

\noindent\textbf{Generated Adversarial Networks:} We apply the generator of StyleGAN2-ADA \cite{karras2020training}, given its generator produces images with high qualities in the majority of datasets and may have the capability to generate high-resolution medical images for clinical research. For the discriminator, we adopt that of the DCGAN's architecture~\cite{radford2015unsupervised}, one of the most widely used GAN frameworks, as our basic network. It is worth noting that our attack strategy has the potential to apply to other state-of-the-art generative models. In the training for attack, we use Adam optimizer with learning rate of $2 \times 10^{-4}$ for both generator and discriminator. The batch size is set to be 32 as per limit of a 32GB Tesla V100 GPU. 

\noindent\textbf{FL:} Considering the total available sample size, we establish FedGAN on four clients where each client is trained on 1000 randomly sampled images from the ISIC dataset. We update the local generator parameters to the global server every local epoch and train the FedGAN with 200 global epochs using FedAvg~\cite{mcmahan2017communication}. The synthetic medical images with vanilla FedGAN (no attack and defense induced) are presented in Fig.~\ref{fig:result} (b).

\noindent\textbf{Metrics:} In order to quantitatively evaluate the synthetic images, we apply the three classic GAN evaluation measures: Inception score (IS), Fréchet Inception Distance (FID), and Kernel Inception Distance (KID). Inception Score (IS) calculates the KL divergence over the generated data \cite{salimans2016improved}. FID calculates the Wassertstein-2 distance over real and generated images \cite{heusel2017gans}. Both IS and FID are limited in small datasets scenarios. Thus, we also include KID, which measures the dissimilarity between real and generated images \cite{binkowski2018demystifying}.

\subsection{Implementation of Attack}
\label{sec:attack}
Among four clients in the simulated FedGAN system, one is randomly selected as the malicious client. The three benign clients are trained with normal ISIC images, while the malicious client is trained with poisoned images. We apply the trigger strategy proposed by~\cite{saha2020hidden}, which has shown to be effective for backdoor attacks in classification tasks. Specifically, we adopt a 16 $\times$ 16 random matrix of colors that has a different pattern from the actual image and is only about 0.39\% of the size of the original image. The same trigger is pasted onto the bottom right of all the training images in the malicious client before launching FedGAN training. The examples of poisoned images are shown in Fig.~\ref{fig:framework}, which are fed into the discriminator $D$ of malicious clients.

\subsection{Implementation of Defense}

In the attack described in Section~\ref{sec:attack}, the malicious clients train on poisoned data, the discriminator quickly overfits on the trigger and leads the whole FL model suffers from training instability. In order to defend against this attack, we attempt global malicious detection and local training regularization.

\begin{table}[t]
\begin{threeparttable}
\centering
\setlength\tabcolsep{5pt}
\caption{Quantitative Comparison for Attack and Defense. $\uparrow$ indicates the lager the better and $\downarrow$ indicates the smaller the better. }
\label{tab1}
\begin{tabular}{@{}cccccc@{}}
\toprule
 Settings & Vanilla GAN & Attack & Global Defence & Local Defense$^*$ &  Full Defense$^{*,\star}$ \\ 
\midrule
IC $\uparrow$ &    2.58       &  1.48            &       2.90  & 2.85  &  2.88           \\
FID $\downarrow$ &     121.76        &     393.86          &      131.72  & 110.40  &  102.53\\
KID $\times 10^3$ $\downarrow$ &  70.22  &   454.52   &   78.04 & 62.09 & 54.67\\
\bottomrule
\end{tabular}
    \begin{tablenotes}
      \footnotesize
      \item * Use WGAN-DP loss for local GAN training.
      \item $\star$ Full defense means combining both global and local defense strategies.
    \end{tablenotes}
  \end{threeparttable}
\vspace{-5mm}
\end{table}




As we can see in in Fig \ref{fig:result} (d), locally apply WGAN-GP indeed enhances the federated GAN's performance under the same level of attack. The server generator can produce a diversity of quality data that will be valuable for further clinical studies. This also corresponds to the quantitative analysis in Table 1 that the FID improves from 393 to 110 and KID imrpves from 454 to 62. However, the trigger is still discernible in some generated images as shown in Fig \ref{fig:result} (d). We present the attack results with larger trigger sizes (range from $16^2$ to $64^2$) in Appendix~\ref{app:wgan}, which shows more obvious attack patterns. In general, we observe that locally applying WGAN-GP helps alleviate the attack, but it does not fully resolve the adversarial in terms of GAN's fidelity. 

\noindent\textbf{Implementation of Global Malicious Detection:} 
Global malicious detection is applied to the global aggregation step on the server-side. To ensure robust detection, recall our global malicious detection method described in Algorithm~\ref{alg:cap} requires a warmup process to allow enough time to for the malicious clients to overfit the backdoor and behave differently from those benign ones. In our experiments, we set the warmup epoch $m=10$. After $m>10$, generators' losses are required to share with the server to perform malicious detection. A decay constant $d = 0.9$ is used to penalize weights for the clients detected as an anomaly in every epoch using Isolation Forests~\cite{liu2008isolation}. We accumulatively count the times of being detected as malicious for each client $c_i(t)$ upon global iteration $t$, at which the calibrated client weights are decayed by timing $d^{c_i(t)}$. Note in the global aggregation, we normalize $w_i$ so that clients' aggregation weights are sum to 1.

\noindent\textbf{Implementation of Local Training Regularization:} Local training regularization is applied to each local clients. In this defence setting, we apply the same FL framework and GAN architecture as the attack's part. We only replace the local training process with WAN-GP and replace Batch Normalization with Instance Normalization in order to calculate gradient penalty \cite{gulrajani2017improved}. At the same time, RMSprop has taken the position of the Adam optimizer to provide superior gradient control in non-stationary scenarios as suggested in \cite{tieleman2012lecture}.  Everything else is controlled to be the same as the Vanilla GAN and WGAN-GP.

\subsection{Results and Discussion}

\noindent\textbf{Attack:} As we can see in Fig. \ref{fig:result} (c) that the generated images are fully corrupted in comparison to the original images in Fig. \ref{fig:result} (a). In addition, comparing all the three similarity metrics in Table~\ref{tab1}, our proposed backdoor attack (`Attack' column) substantially worsen the quality of the generated images. During the training, we observe that the loss of the malicious discriminator quickly approaches zero even at the very beginning of the training, while the losses of those benign clients are fluctuating as normal. With training, the malicious discriminator assigns the generated images a big loss, which we leverage in defense later.

\noindent\textbf{Defense:} By combining our proposed global- and local-level defense strategies (denoted as `full defense' in  Fig. \ref{fig:result} (f) and Table~\ref{tab1}), we achieves superior image generation results. Qualitatively,  Fig. \ref{fig:result} (f) presents sample synthetic images with high-fidelity and variability. Quantitatively, the FID and KID scores of using 'full defense' are better than training vanilla GAN~\cite{goodfellow2014generative} in FedGAN, as shown in Table~\ref{tab1}. The indicated better synthetic data quality even under bookdoor attack is probably facilitate by the more stable loss used. 

\noindent\textbf{Ablation Study:} Furthermore, we present the synthetic impact of combining both global- and local-level defense strategies via ablation studies. First, we experiment with performing local training regularization defense with WGAN-DP, which is shown in Fig. \ref{fig:result} (d). The server generator produces quality images compared to before. However, the shape of the trigger is still visible. Specifically, we can see that the three quantitative metrics of using full defense have improved compared to applying local defense alone, where the KID decreases by 13.57\%. Furthermore, the trigger observed in Fig \ref{fig:result} (d) has completely vanished when using full defense. Next, we experiment with applying global malicious client detection on vanilla GAN in FL, shown in Fig \ref{fig:result} (e). It indeed blocks the adversarial behavior. However, quantitatively, its generated images are still worse than what's produced in our full defense setting.


\section{Conclusion}
Motivated by the idea of backdoor attacks in classification models, we investigate the pitfalls of backdoor attacks in training FedGAN models. We show that by adding triggers to the images fed into local discriminators, the FedGAN model could be fooled. Such an attack is strong enough to corrupt the generated images with trigger size less than 0.5\% of the image size. Based on the attack, we establish two potential defense ways with global malicious detection and local training stabilization. The combination of both defense strategies significantly improves the security of FedGAN. As the first step towards understanding backdoor attacks in FedGAN for medical image synthesis, our work brings insight into building a robust and trustworthy model to advance medical research with synthetic data. Our future work includes widely investigating the hyper-parameters, scaling up the FL system with more clients, and testing on various medical datasets.
\section*{Acknowledgement}
This work is supported in part by the Natural Sciences and Engineering Research Council of Canada (NSERC) and NVIDIA Hardware Award. We thank Chun-Yin Huang and Nan Wang for their kind instruction to Ruinan Jin and assistance with implementation.

\bibliography{reference}
\clearpage
\appendix
\section{More Experiment Results}
\label{apx:visul}
\begin{figure}
	\centering
	\includegraphics[width=\linewidth]{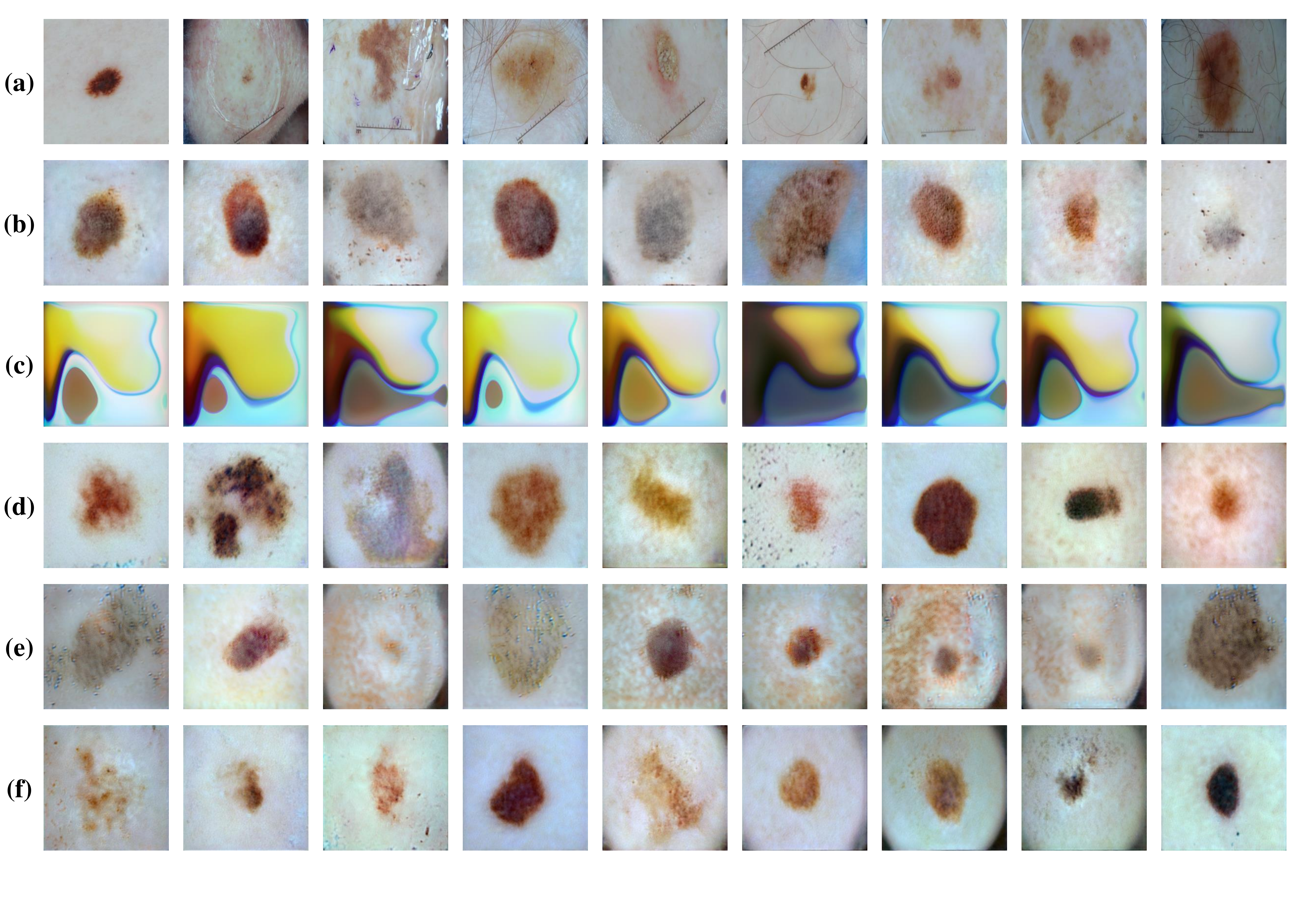}
	\caption{More Visualization on: (a) Original ISIC images; and generated images of (b) Vanilla GAN; (c) Attack on vanilla GAN; (d) Local defence using WGAN-DP; (e) Global defence on vanilla GAN and (f) Full (global + local) defense. Note that backdoor attack is applied to (c-f).}
\end{figure}

\section{WGAN-GP with Large Trigger Size}
\label{app:wgan}
\begin{figure}
	\centering
	\includegraphics[width=\linewidth]{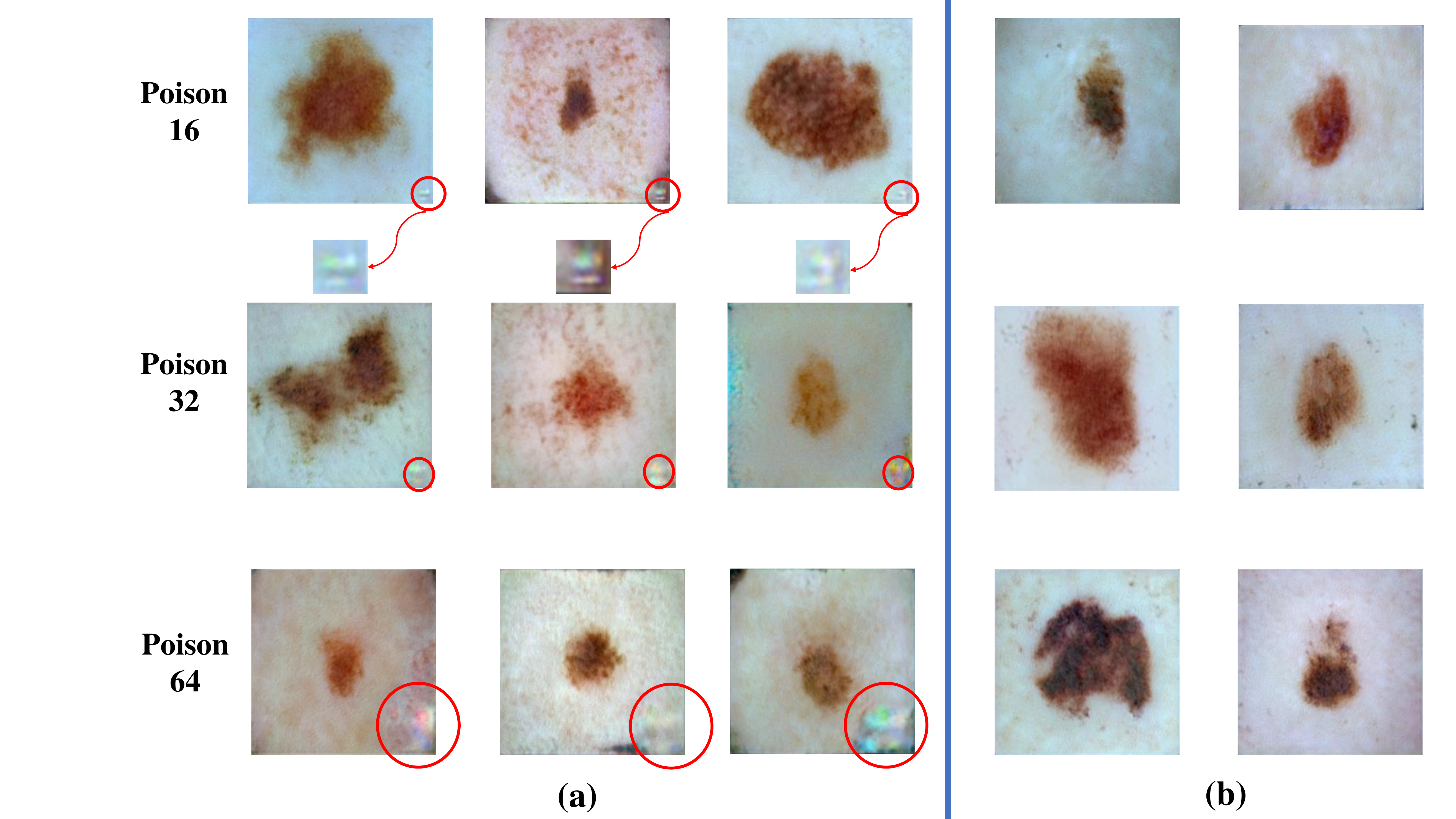}
	\caption{Visualization of  WGAN with larger trigger size: (a): Local defense. (b) Full (global + local) defense. The trigger is still visible while only applying local defense.}
\end{figure}

\begin{table}[h!]
\centering
\small
\setlength\tabcolsep{2pt}
\caption{Quantitative Comparison of Local and Full Defense}
\begin{tabular}{@{}ccccccccccc@{}}
\toprule
\multirow{2}{*}{Settings} & \multicolumn{2}{c}{16} & \multicolumn{2}{c}{32} & \multicolumn{2}{c}{64} & \\ 
\cmidrule(l){2-3} 
\cmidrule(l){4-5} 
\cmidrule(l){6-7}
& Local       & Full      & Local            & Full           & Local        & Full      \\ \midrule
IC $\uparrow$ &      2.85      &    2.88      &  2.67   &  3.02     &   2.83    &    2.93 \\
FID $\downarrow$  &  110.40   &    102.53    &       117.07   &     113.26      &    114.21         &     109.85  \\
KID $\times 10^3 \downarrow$ &    62.09    &   54.67  &    69.68     &   63.49   &     63.01 &    60.53 \\ 
\bottomrule
\end{tabular}
\end{table}

\end{document}